\documentclass[10pt,twocolumn,letterpaper]{article}

\usepackage{cvpr}              

\usepackage{mdwmath}
\usepackage{eqparbox}
\usepackage{epsfig}
\usepackage{graphicx}
\usepackage{amsmath}
\usepackage{amssymb}
\usepackage{pifont}
\usepackage{url}            
\usepackage{booktabs}       
\usepackage{tabu}           
\usepackage{multirow}
\usepackage{graphicx}
\usepackage{algorithm}
\usepackage{algorithmicx}
\usepackage{algpseudocode}
\usepackage{array}

\usepackage{cite}

\usepackage{makecell}
\usepackage{tabularx}
\usepackage{pifont}
\usepackage{multicol}
\usepackage{array}
\usepackage{adjustbox}

\usepackage{url}            
\usepackage{booktabs}       
\usepackage{amsfonts}       
\usepackage{nicefrac}       
\usepackage{microtype}      
\usepackage{graphicx}
\usepackage{amsmath}
\usepackage{bm}
\usepackage{multirow}
\usepackage{tabu}
\usepackage{siunitx}
\usepackage{adjustbox}
\usepackage{subcaption}
\usepackage{siunitx}
\usepackage{colortbl}
\usepackage{color}
\usepackage{pifont}
\definecolor{Gray}{gray}{0.9}
\definecolor{Lightorange}{RGB}{255,214,169}
\definecolor{Cyan}{rgb}{0.88,1,1}
\definecolor{reminder}{RGB}{255,0,0}

\usepackage{tabu}           
\usepackage{multirow}
\usepackage{booktabs}
\usepackage{makecell}
\usepackage{pifont}
\usepackage{longtable}
\usepackage{wrapfig}
\usepackage{subcaption}

\newcolumntype{x}[1]{>{\centering\arraybackslashå}p{#1pt}}
\newlength\savewidth

\newcommand{\PreserveBackslash}[1]{\let\temp=\\#1\let\\=\temp}
\newcolumntype{C}[1]{>{\PreserveBackslash\centering}p{#1}}
\newcolumntype{L}[1]{>{\PreserveBackslash\raggedright}p{#1}}

\newcommand{\benchname}{Demo-ICL-Bench\xspace}
\newcommand{\modelname}{Demo-ICL\xspace}
\newcommand{\taskname}{Demo-driven Video In-Context Learning\xspace}
\newcommand{\taskabbr}{Demo-driven ICL\xspace}
\definecolor{green1}{RGB}{0, 176, 80}
\definecolor{red1}{RGB}{192, 0, 0}

\usepackage{url}
\usepackage{marvosym}

\definecolor{cvprblue}{rgb}{0.21,0.49,0.74}
\usepackage[pagebackref,breaklinks,colorlinks,allcolors=cvprblue]{hyperref}

\def\paperID{} 
\def\confName{CVPR}
\def\confYear{2026}

\title{Demo-ICL: In-Context Learning for Procedural Video Knowledge Acquisition}

\author{Yuhao Dong$^1$\thanks{Authors contributed equally to this research.~~\textsuperscript{~\Letter}Corresponding authors.}\quad 
Shulin Tian$^1$\footnotemark[1]\quad 
Shuai Liu$^1$\quad
Shuangrui Ding$^{2,3}$\quad \\
Yuhang Zang$^2$\textsuperscript{~\Letter}\quad
Xiaoyi Dong$^2$\quad
Yuhang Cao$^2$\quad
Jiaqi Wang$^2$\quad
{Ziwei Liu}$^1$\textsuperscript{~\Letter}\\
$^1$ S-Lab, Nanyang Technological University \quad
$^2$ Shanghai AI Lab \quad
$^3$ CUHK-MMLab \quad \\
\url{https://github.com/dongyh20/Demo-ICL}
}

\begin{document}
\twocolumn[{
    \renewcommand\twocolumn[1][]{#1}
    \maketitle
    \begin{center}
        \includegraphics[width=1.0\linewidth]{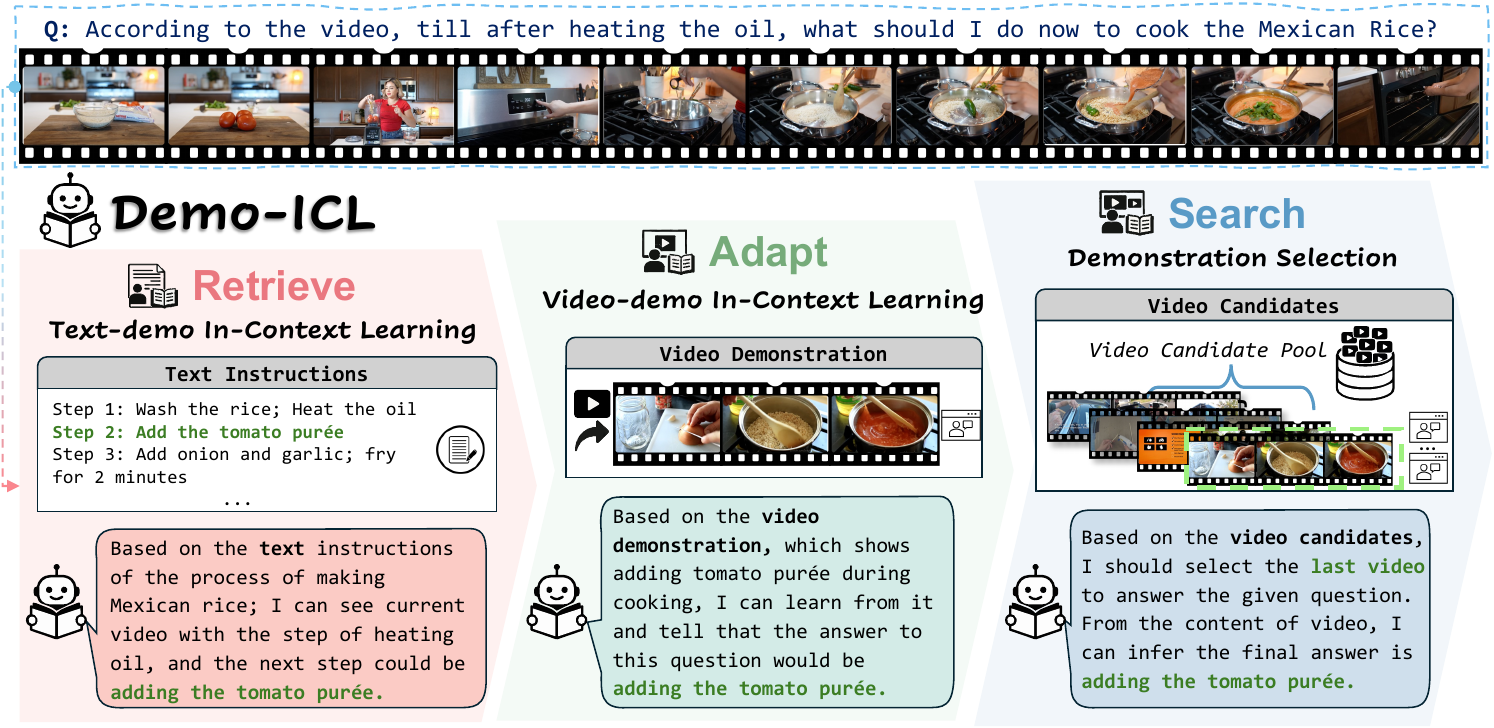}\vspace{-5pt}
        \vspace{-6pt}
        \captionof{figure}{\textbf{Overview of the \taskname Task} with three distinct settings: (1) \textit{Text-demo in-context learning}, where text instructions act as the demonstrations; (2) \textit{Video-demo in-context learning}, where a video demonstration acts as the reference; and (3) \textit{Demonstration Selection}, which requires identifying the most relevant video demonstrations among the video candidate pool and using them to guide video in-context learning.}
        \label{fig:teaser}
    \end{center}
}]

\renewcommand{\thefootnote}{\fnsymbol{footnote}} 
\footnotetext[1]{Authors contributed equally to this research. \textsuperscript{\Letter} Corresponding authors.} 
\renewcommand{\thefootnote}{\arabic{footnote}} 
\setcounter{footnote}{0} 

\begin{abstract}
Despite the growing video understanding capabilities of recent Multimodal Large Language Models (MLLMs), existing video benchmarks primarily assess understanding based on models' static, internal knowledge, rather than their ability to learn and adapt from dynamic, novel contexts from few examples.
To bridge this gap, we present \textbf{\taskname}, a novel task focused on learning from in-context demonstrations to answer questions about the target videos.
Alongside this, we propose \textbf{\benchname}, a challenging benchmark designed to evaluate demo-driven video in-context learning capabilities.
\benchname is constructed from 1200 instructional YouTube videos with associated questions, from which two types of demonstrations are derived: (i) summarizing video subtitles for text demonstration; and (ii) corresponding instructional videos as video demonstrations.
To effectively tackle this new challenge, we develop \textbf{\modelname}, an MLLM with a two-stage training strategy: video-supervised fine-tuning and information-assisted direct preference optimization, jointly enhancing the model's ability to learn from in-context examples.
Extensive experiments with state-of-the-art MLLMs confirm the difficulty of \benchname, demonstrate the effectiveness of \modelname, and thereby unveil future research directions.
\end{abstract}    
\vspace{-30pt}
\section{Introduction}
\label{sec:intro}

Video understanding remains challenging.
Recent Multimodal Large Language Models (MLLMs) \citep{alayrac2022flamingo, awadalla2023openflamingo,huang2023language,zhao2023mmicl,peng2023kosmos,wang2024qwen2vlenhancingvisionlanguagemodels,internvl,li2024llavaonevisioneasyvisualtask,dong2025insight} show progress from short-clip recognition \citep{caba2015activitynet,mangalam2023egoschemadiagnosticbenchmarklongform,goyal2017somethingsomethingvideodatabase,zhang2025towards} to long videos analysis \citep{chandrasegaran2024hourvideo} from daily-life settings \citep{lei2019tvqalocalizedcompositionalvideo,mangalam2023egoschemadiagnosticbenchmarklongform} to instructional videos \citep{miech2019howto100m,tang2019coinlargescaledatasetcomprehensive,zhukov2019cross}.
However, existing video benchmarks typically pose questions that rely on either internal pre-trained knowledge (e.g., asking ``what is a whisk?'') or by visible facts in the target video (e.g., ``where is the whisk?'').
This is fundamentally different from a more challenging scenario where a model must learn a new process or skill from demonstrations (e.g., a video tutorial that teaches the model how to cook Mexican Rice) and then apply that learned knowledge to answer questions based on a new, related target video sequence. This scenario reflects human learning and is crucial for downstream applications like robotics, where robots can learn from demonstrations to tackle new tasks.
For example, in \cref{fig:teaser}, after seeing only the first step of heating oil for Mexican Rice, the model is asked ``what should you do next?'' based on \textit{in-context} text instructions or video demonstrations.
This question requires knowing the specific sequence of steps for this particular version of Mexican Rice that the model is presumably meant to understand or follow, such as based on the in-context video demonstrations.

To better encourage models to learn new skills from context and adapt to novel tasks, we propose a challenging video understanding task called \textbf{\taskname(\taskabbr)}.
Our task embodies this by presenting target videos and questions alongside \textit{in-context} text guidelines or video demonstrations.
As shown in \cref{fig:teaser}, \taskabbr has three sub-settings: (1) text-demo in-context learning, (2) video-demo in-context learning, and (3) demonstration selection.
These questions explicitly require models to use the knowledge provided within in-context examples, rather than just their static internal knowledge.
A key difference between \taskabbr and previous in-context learning paradigms lies in the data modality and interaction type: our in-context demonstrations can be videos, and the task involves choosing from a candidate pool to identify suitable in-context examples by the model itself.
The \taskabbr tasks mirror how one might learn a complex skill, such as cooking, by searching for related video demonstrations and watching them while also consulting supplementary visual or textual guides.

To evaluate the proposed Demo-driven ICL task, we present \textbf{\benchname}, a benchmark that consists of text/video demonstrations, target videos, questions, and answers.
We collected instructional YouTube videos from the HowTo100M dataset \citep{miech2019howto100m}, ensuring subtitles and timestamps.
Subsequently, we used an LLM to summarize these subtitles, generating text demonstrations to serve as in-context examples.
Additionally, we employ video search ranking methods and an LLM to identify and select videos similar to the target video to serve as in-context video demonstrations, and we also construct a video candidate pool for the model to select from and learn, in order to mimic real-world scenarios.
\benchname is complex and challenging: answering every question demands an accurate understanding of the demonstrations, resulting in frontier models like Gemini-2.5-Pro achieving merely 46.6\% and 32.0\% accuracy when processing text and video demonstrations, respectively.

To address \taskabbr, we present \textbf{\modelname} with a two-stage training strategy: video supervised fine-tuning, and information-assisted Direct Preference Optimization (DPO) for demo-driven video ICL.
We design an information-assisted DPO data generation pipeline to produce high-quality chosen responses by simplifying the task with additional contextual information.
\modelname outperforms existing MLLMs on the proposed \taskabbr, VideoMMMU for video knowledge acquisition \citep{hu2025videommmuevaluatingknowledgeacquisition}, and VideoMME~\citep{videomme} for general video understanding.

Our key contributions are: \textbf{(i)} \textbf{New Challenging Tasks}: We design three \taskname (\taskabbr) tasks, which enable models to answer questions by learning from text or video demonstrations, representing a significant step towards more human-like learning and decision-making processes in video understanding tasks.
\textbf{(ii)} \textbf{New Benchmark and Evaluation}: We establish a new \benchname that is specifically designed for evaluating demo-driven video in-context learning capabilities.
Based on \benchname, we conduct comprehensive evaluations of cutting-edge baselines, showcasing various challenges of our proposed task.
\textbf{(iii)} \textbf{New \modelname Model}: We present a new model, \modelname, along with a customized two-stage training strategy that enhances a model's ability to learn and adapt from in-context demonstrations.
Compared with SOTA models, \modelname shows competitive performance across existing benchmarks \citep{videomme,wu2024longvideobench,hu2025videommmuevaluatingknowledgeacquisition}, demonstrating its superior video comprehension and in-context knowledge acquisition capabilities.

\section{Related Work}
\label{sec:related_work}

\noindent \textbf{Multimodal Video Understanding for Knowledge Acquisition.}
Multimodal video understanding is evolving from low-level perception toward \textit{knowledge acquisition} - the ability to extract, structure, and apply information from complex instructional videos, where large-scale instructional datasets play a pivotal role in this shift. HowTo100M~\citep{miech2019howto100m} introduced 1.2 million narrated videos with 136 million clip–caption pairs for procedure recognition and cross-task transfer. Other instruction-based datasets~\citep{tang2019coinlargescaledatasetcomprehensive,zhukov2019cross} provide fine-grained task annotations or support weakly supervised step parsing across diverse procedures. More recently, benchmarks like Video-MMMU~\citep{hu2025videommmuevaluatingknowledgeacquisition}, Video-MMLU~\citep{song2025video}, and VideoMathQA~\citep{rasheed2025videomathqa} start to probe models' capabilities to learn from educational videos, shifting emphasis from perception to knowledge uptake and application.
To further benchmark the video understanding task for knowledge acquisition that is closer to real-world settings, we introduce \taskname and propose \benchname, to systematically evaluate models' capabilities of acquiring new concepts via given video demonstrations.

\noindent \textbf{Multimodal In-Context Learning.}
In-context learning (ICL) enables models to perform new tasks by conditioning on a few examples at inference. Initially developed for large language models (LLMs), ICL has been extended to multimodal settings and shows consistent performance gains across language and image tasks \citep{brown2020languagemodelsfewshotlearners,min2021metaicl,luo2024context,zhou2024visual}. However, video-based ICL remains underexplored: current video MLLMs~\citep{video-llava,maaz2024videochatgptdetailedvideounderstanding,zhang2025llavavideovideoinstructiontuning} mainly emphasize zero-shot performance through curated video instruction datasets for open-ended QA, captioning, and dialog capabilities. Emerging works add chain-of-thought methods for video understanding and reasoning tasks~\citep{wang2024videocot,han2025videoespresso,arnab2025temporalchainthoughtlongvideo,tian2025ego,zhang2025vitcot,ghazanfari2025chain} encourage stepwise evidence aggregation and explicit explanation, while some retrieval-based methods like VideoRAG~\citep{tevissen2024towards,ren2025videorag} establish a new paradigm of retrieving video moments and ground answers in cited segments. However, the aforementioned works merely use the context as a reference, rather than adapting to and learning from the provided context. In our work, we address this gap by introducing the \taskabbr task on instructional video datasets, supported by an optimized training pipeline, to enhance the model's capability to learn from the in-context video demonstrations.
\section{Demo-ICL: Procedural Knowledge Learning from In-Context Demonstrations}
In this section, we provide a detailed task formation and dataset construction of the proposed \taskname tasks.
\cref{sec:task} presents task definitations, \cref{sec:construction} outlines the dataset construction process. 
Finally, we demonstrate how we train our model to achieve demo-driven video knowledge acquisition in \cref{sec:training}.

\begin{figure*}
    \centering
    \includegraphics[width=\linewidth]{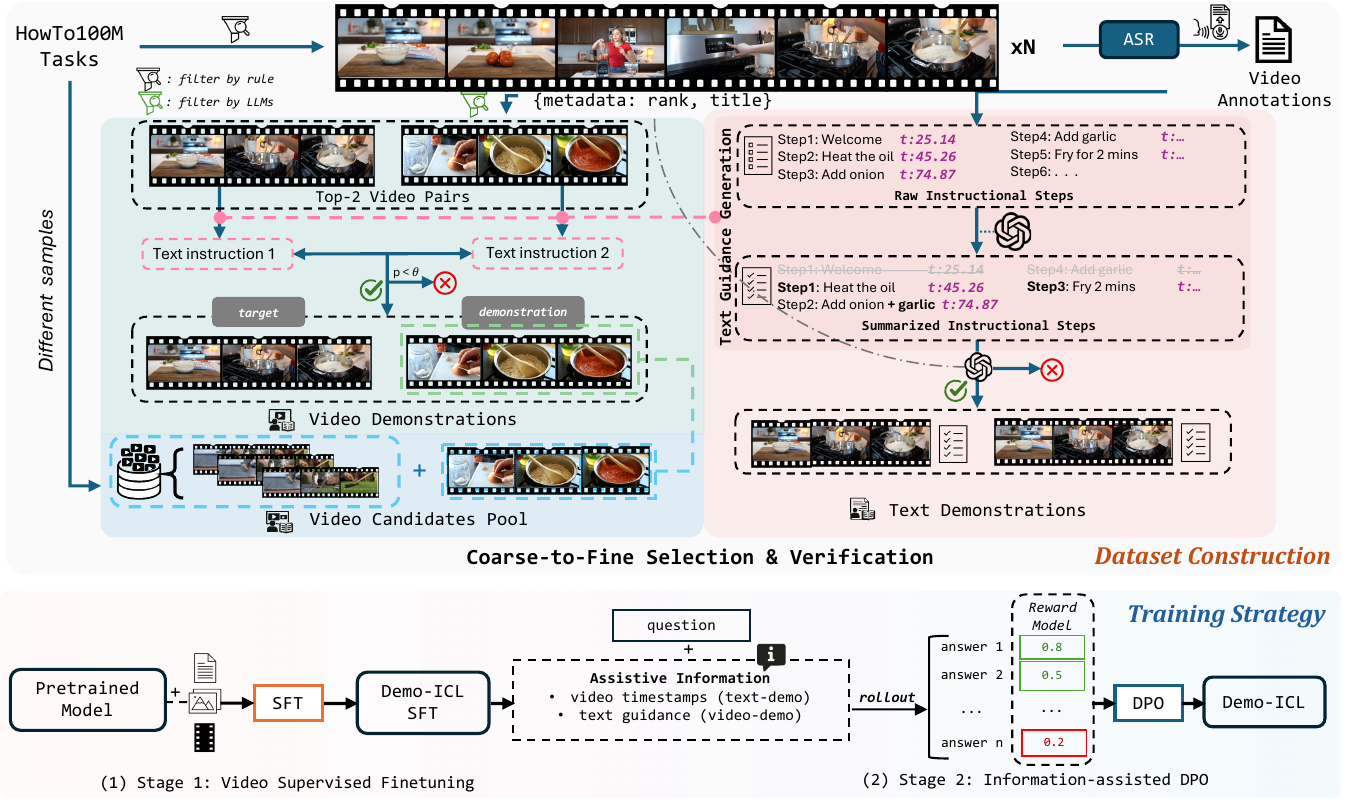 }
    \vspace{-10pt}
    \caption{\textbf{Overview of Data Construction and Training Strategy.} (i) illustrates our coarse-to-fine dataset collection pipeline (\cref{sec:construction}); (ii) presents the tailored two-stage training strategy for training the \modelname model (\cref{sec:training}).}
    \label{fig:data}
    \vspace{-10pt}
\end{figure*}
\subsection{Demo-driven Video In-Context Learning}
\label{sec:task}
Learning from demonstrations and imitating actions are crucial skills for humans when acquiring new abilities. Such capabilities enable individuals to rapidly master novel tasks from only a handful of examples, thereby supporting efficient adaptation and facilitating lifelong learning. In contrast, contemporary video models largely depend on supervised fine-tuning to acquire task-specific capabilities, neglecting the importance of learning from in-context examples and evaluating such capabilities for achieving human-like performance. Additionally, humans often learn new tasks incrementally. This alignment underscores the need for models that support procedural video knowledge acquisition, enabling the incremental internalization and generalization of task procedures in a human-like manner.

To address such problems, we propose a new set of three tasks called \textbf{\taskname}. These three tasks are designed to evaluate the model's ability to learn from in-context demonstrations. Given an instructional video $V_D$ or text demonstration $T_D$, the model must first interpret the example to understand how a task should be completed. It is then presented with a test video $V_{Test}$, and its ability to transfer knowledge from the demonstration is assessed by predicting subsequent steps of an action $A_{[t_1, t_2]}$ based on the demonstration and the available context $V_{Test}[0,t_1]$. Depending on the format and source of demonstrations, we define three distinct tasks:

\textbf{1) Text-demo In-Context Learning:} The model answers questions from an input video (e.g., ``what should I do now to cook Mexican Rice?'') by retrieving information from the corresponding \textbf{textual instructions} (e.g., ``Step 1: Wash the rice; Step 2: Add the tomato; Step 3: \ldots''), which serve as the demonstration.
For example: ``Based on the \textit{text instructions}, the current video corresponds to Step 2. Therefore, the next step is Step 3: add the onion.''

\textbf{2) Video-demo In-Context Learning}: The model answers questions about a target video by conditioning on a provided \textbf{video demonstration} of a similar task, and using the demonstration as an in-context exemplar.
For instance: ``Given the \textit{video demonstration}, the target clip aligns with Step 2; therefore, the next step is Step 3: add the onion.''

\textbf{3) Demonstration Selection}: The model is given the input video and a pool of video candidates (e.g., a pool containing ``Mexican rice,'' ``fried rice,'' and ``pasta,'').
The model must first \textbf{select the most relevant demonstration} (e.g., ``Mexican rice'') from the pool and then use it to answer the question, simulating a scenario where a perfectly aligned demonstration is not provided.

Collectively, these three tasks constitute a systematic and comprehensive framework for demo-driven video in-context learning. They underscore distinct model capabilities, ranging from textual retrieval to demo-based knowledge extraction and adaptation, and correspond to successive stages of development, spanning from idealized oracle settings to practical real-world scenarios.

\subsection{Dataset Construction}
\label{sec:construction}
In this section, we introduce a comprehensive data generation pipeline to support the proposed demo-driven video ICL task. The pipeline emphasizes four key qualities: ensuring that the video content is informative, the textual demonstration is precise, the video demonstrations are contextually relevant, and the generated questions remain answerable. We develop a structured step-by-step process (see \cref{fig:data} (i)) and detail each component below.

\noindent \textbf{Video Collection and Annotation.}
We use video data from HowTo100M~\citep{miech2019howto100m}, a large-scale corpus of narrated YouTube instructional videos designed for complex tasks. This dataset is particularly suitable for demo-driven video in-context learning, as it allows models to acquire procedural knowledge from step-by-step demonstrations. With over 100 million clips covering 23,000 activities, HowTo100M provides diverse and extensive instructional material for building our benchmark. After selecting this source, we first filter videos based on video length, language, and title availability. To obtain high-quality annotations, we use ASR outputs as they offer more detailed descriptions of demonstrated activities compared with video captions. Specifically, we use annotations from HTM-AA~\citep{han2022temporal}, which employs WhisperX~\citep{bain2023whisperx} to generate sentence and word-level timestamps. 

\noindent \textbf{Text Demonstration Generation.}
We generate textual demonstrations for each video using a coarse-to-fine pipeline that produces step-by-step instructions. First, we use Qwen2.5-72B~\citep{bai2025qwen25} to summarize the ASR transcripts into a sequence of clips, identifying the instructional steps needed to complete the task. Next, we filter out irrelevant steps, retaining valid steps while merging redundant ones into the nearest relevant segment to preserve continuity. This process yields a coherent, task-focused sequence of instructions. Finally, we incorporate Qwen2.5-VL-72B~\citep{bai2025qwen25} to refine the demonstrations by jointly considering the step descriptions and corresponding video clips, ensuring contextual accuracy and close alignment with the depicted actions. Through this multi-stage refinement, we obtain precise and reliable textual guidance that captures both the procedural structure of the task and its visual grounding.

\noindent \textbf{Video Demonstration Selection.}
To enable video-based in-context learning, we construct pairs of videos that illustrate similar tasks. These paired demonstrations serve as explicit visual guidance, allowing the model to observe alternative executions and acquire procedural knowledge from reliable examples. Pair selection follows a coarse-to-fine process. We first leverage metadata from HowTo100M, which provides YouTube search rankings for specific tasks, and discard tasks with few relevant videos to ensure pairing quality. Next, we evaluate the titles of top-ranked videos for each task and select the two with the highest semantic similarity using Qwen2.5-72B. To further validate these candidates, we generate textual instructions for both videos using the previous pipeline, after which the LLM compares the instructions to confirm that the videos indeed demonstrate similar tasks that can be transferred. This process ensures both accuracy and reliability of the selected video guidance.

\noindent \textbf{Demo-driven Question Generation.} 
With curated videos and corresponding instructions, we construct questions to evaluate demo-driven video in-context learning. For text-based in-context learning, we exclude tasks with fewer than six steps to ensure sufficient complexity. From each valid sequence, one intermediate step (excluding the first and last) is randomly selected, and a question is generated from this step. The model is then required to predict the next action.

For video-based in-context learning, the goal is to test whether models can effectively leverage visual demonstrations. To this end, we employ an LLM to analyze the generated instructions for paired videos, determining whether they represent comparable tasks suitable for question generation. If validated, the LLM identifies the target step and its corresponding timestamp within the pair. Human annotators then assess the generated questions, focusing on whether the visual demonstrations provide meaningful contextual evidence for answering. To avoid trivial cases, we further filter out highly similar video pairs, ensuring the task meaningfully tests the model’s adaptability and generalization.

For the demonstration selection task, the validated video pairs are treated as ground truth and augmented with 3 carefully chosen irrelevant videos. To ensure diversity, for each video, we dynamically select the video candidates to form a unique video candidate pool for each question, rather than keeping the pool static. This construction requires models to distinguish informative demonstrations from distractors, an essential capability for real-world deployment.

\noindent \textbf{Dataset Partition and Benchmark Statistics.} Following established protocols, we first generate 5,000/2,000/1,000 questions for text-demo ICL/video-demo ICL and demonstration selection settings. To construct \benchname, we then manually curate representative video demonstrations that highlight the role of demo-driven video in-context learning. Specifically, we sample 500 questions each for the text-demo and video-demo settings and 200 questions for the demonstration selection setting, resulting in a balanced benchmark of 1,200 questions in total.

\subsection{Learning from In-Context Demonstrations}
\label{sec:training}
We further train Demo-ICL models to validate the effectiveness of the proposed demo-driven video in-context learning task. The training pipeline is intentionally simple yet effective. As shown in \cref{fig:data} (ii), we adopt a two-stage strategy to progressively integrate demo-driven in-context learning. The model is first fine-tuned on a tailored dataset to enhance fine-grained video comprehension and general in-context reasoning. Then we employ a customized DPO framework to specifically strengthen the model’s capacity to learn from video demonstrations in context.

\subsubsection{Video Supervised Fine-tuning} 
In this stage, our goal is to equip Demo-ICL with fine-grained video understanding and general in-context reasoning capabilities. To this end, we compile a large-scale dataset containing millions of samples drawn from diverse text–image pairs and video sources in open academic repositories. For image–text data, we rely on resources such as LLaVA-OneVision \citep{li2024llavaonevisioneasyvisualtask}, VisualWebInstruct \citep{jia2025visualwebinstruct}, and other widely used collections. For video data, we incorporate material from open-source projects including LLaVA-Video \citep{zhang2024video}, Oryx \citep{liu2024oryx}, and Ola \citep{liu2025olapushingfrontiersomnimodal}. To further enhance the model’s ability for instructional video understanding, we additionally incorporate datasets such as COIN \citep{tang2019coinlargescaledatasetcomprehensive} and Cross-Task \citep{zhukov2019cross}. We carefully exclude any videos that overlap with \benchname to prevent data leakage and ensure fair evaluation. Finally, we perform subsampling on the generated dataset as described in \cref{sec:construction} to explicitly introduce demo-driven video in-context learning signals during this stage. Together, these curated resources establish robust foundational capabilities and prepare the model for subsequent stages to enhance demo-driven video in-context learning.

\subsubsection{Information-Assisted Preference Optimization}

Preference learning has become a critical component in the advancement of large language models, aiming to fine‑tune outputs to better align with human preferences and improve real‑world applicability. Traditional DPO algorithms generate multiple responses to the same query, after which a reward model ranks them and identifies preferred and rejected responses for training.
\begin{table*}[t]
\centering
\caption{\textbf{Evaluation results on \benchname.} The benchmark assesses models across three tasks. For Text-demo ICL and Video-demo ICL, we report two types of accuracy: Demo. Acc (with demos) and w/o Demo(without demos), and the improvement (\boldmath$\Delta_{\text{ICL}}$) attributed to demonstrations. DS refers to Demonstration Selection task. S.Acc refers to demonstration selection accuracy.
}\vspace{-5pt}
\label{tab:iclresult}
\adjustbox{width=\linewidth}{
\begin{tabular}{@{}l c c|c c c |c c c |c c|c@{}}
\toprule
\multirow{2}{*}{\textbf{Model}} &
\multirow{2}{*}{\textbf{Size}} &
\multirow{2}{*}{\textbf{Frame}} &
\multicolumn{3}{c|}{\textbf{Text-demo ICL}} &
\multicolumn{3}{c|}{\textbf{Video-demo ICL}} &
\multicolumn{2}{c|}{\textbf{DS}} &
\multirow{2}{*}{\textbf{Avg}} \\
\cmidrule(lr){4-6} \cmidrule(lr){7-9} \cmidrule(lr){10-11}
& & & 
Demo. Acc & w/o Demo. & $\Delta_{\text{ICL}}$ &
Demo. Acc & w/o Demo. & $\Delta_{\text{ICL}}$ &
S.Acc & Acc \\
\midrule
Human & -  & - & 84.0 & - & - & 80.4 & - & - & 88.0 & 76.0 & 80.1 \\
\midrule
\textit{Proprietary MLLMs} \\
\midrule
Gemini-2.5-Pro~\citep{gemini2024gemini15} & - & - & 54.4 & - & - & 36.2 & - & - & - & 26.0 & 38.9 \\
GPT-4o~\citep{openai2024gpt4o} & - & -  & 48.8 & - & - & 31.4 & - & - & - & 24.5 & 34.9 \\
\midrule
\textit{Open-Source Video MLLMs} \\
\midrule
Qwen2-VL~\citep{wang2024qwen2vlenhancingvisionlanguagemodels} & 7B & 32  & 29.0 & 21.8 & \colorbox{green1!20}{+7.2} & 22.4 & 24.0 & \colorbox{red1!10}{-1.6} &  38.0 & 14.5 & 22.0 \\
Ola~\citep{liu2025olapushingfrontiersomnimodal} & 7B & 32  & 32.2 & 23.0 & \colorbox{green1!20}{+9.2} & 24.6 & 26.4 & \colorbox{red1!10}{-1.8} & 43.0 & 16.0 & 24.3 \\
LLaVA-Video~\citep{zhang2024video} & 7B & 32  & 31.0 & 25.4 & \colorbox{green1!20}{+5.6} & 30.2 & 29.0 & \colorbox{green1!10}{+1.2} & 44.5 & 20.5 & 27.2 \\
Qwen2.5-VL~\citep{bai2025qwen25} & 7B & 32  & 32.8 & 26.0 & \colorbox{green1!20}{+6.8} & 28.0 & 26.2 & \colorbox{green1!10}{+1.8} & 46.0 & 18.0 & 26.3 \\
InternVL-3~\citep{zhu2025internvl3} & 8B & 32 & 31.4 & 26.6 & \colorbox{green1!10}{+4.8} & 27.0 & 26.4 & \colorbox{green1!10}{+0.6} & 44.0 & 16.5 & 25.0\\
Video-R1~\citep{feng2025video} & 7B & 32 & 33.6 & 27.6 & \colorbox{green1!20}{+6.0} & 27.4 & 26.6 & \colorbox{green1!10}{+0.8} & 48.0 & 17.5 & 26.2\\
VideoChat-R1~\citep{li2025videochat} & 7B & 32 & 34.4 & 27.0 & \colorbox{green1!20}{+7.4} & 28.2 & 26.8 & \colorbox{green1!10}{+1.4} & 52.0 & 18.5 & 27.0\\
\rowcolor{gray!20!white}
Qwen2.5-VL~\citep{bai2025qwen25} & 72B & 32 & 45.0 & 24.2 & \colorbox{green1!30}{+20.8} & 25.6 & 25.2 & \colorbox{green1!10}{+0.4} & 54.0 & 18.0 & 29.5\\
\midrule
\rowcolor{gray!20!white}
Ola-Video~\citep{liu2025olapushingfrontiersomnimodal} (Base) & 7B & 32 & 31.4 & 22.8 & \colorbox{green1!30}{+8.6} & 25.0 & 26.0 & \colorbox{red1!10}{-1.0} & 48.0 & 18.0 & 24.8\\
\rowcolor{blue!10!white}
\modelname (SFT) & 7B & 32 & 38.4 & 27.8 & \colorbox{green1!30}{+10.6} & 29.4 & 26.2 & \colorbox{green1!20}{+3.2} & 54.5& 21.5 & 29.8\\
\rowcolor{blue!20!white}
\modelname & 7B & 32 & \textbf{43.4} & 29.4 & \colorbox{green1!30}{+14.0} & \textbf{32.0} & 27.6 & \colorbox{green1!20}{+4.4} & 58.0 & 24.0 & \textbf{33.1} \\
\bottomrule
\end{tabular}
}
\vspace{-12pt}
\end{table*}

However, current models struggle with demo-driven video in-context learning, limiting their ability to generate high-quality responses. This limitation makes conventional DPO data construction pipelines less effective. To overcome these challenges, we propose an information-assisted DPO pipeline that integrates automatically generated assistive information, eliminating the need for manual annotation. For text-demo ICL tasks, we supply video timestamps to better align visual inputs with textual instructions, thereby improving accuracy. For video-demo ICL tasks, we pair video demonstrations with corresponding textual guidance to enhance response quality. For training, we define a preference dataset as \(\mathcal{P} = \{(x^{(i)}, R_{c}^{(i)}, R_{r}^{(i)})\}_{i=1,\dots,|\mathcal{P}|}\), where each \(x^{(i)}\) denotes the user request, and \(R_{c}^{(i)}\) and \(R_{r}^{(i)}\) represent the preferred and less preferred responses. We employ a reward model \(r^*(x, y)\) to approximate preferences, and a higher score denotes a stronger preference. Following the approach introduced by~\citep{rafailov2024direct}, we can model the human preference distribution \(p^*\) using the Bradley-Terry (BT) model~\citep{bradley1952rank}:
\begin{equation}
\begin{aligned}
&& p^*(y_1 \succ y_2 \mid x) = \frac{\exp(r^*((x,\textit{I}), y_1))}{\exp(r^*((x,\textit{I}), y_1)) + \exp(r^*(x, y_2))} \\
&& = \sigma(r^*((x,\textit{I}), y_1) - r^*(x, y_2)),
\end{aligned}
\end{equation}
where \textit{I} denotes the assistive information, and \(\sigma\) denotes the logistic function. To estimate the parameters of the reward model, we can formulate the problem as a binary classification task and minimize the negative log-likelihood:
\begin{equation}
    \mathcal{L}_R(r_\phi, \mathcal{P}) = - \mathbb{E}_{(x, R_{c}, R_{r}) \sim \mathcal{P}} [\log \sigma(r_\phi(x, R_{c}) - r_\phi(x, R_{r}))],
\end{equation}
where \(r_\phi\) is the reward model. This approach enables effective alignment with human preferences by allowing the model to use additional information that can be generated automatically, thus producing high-quality responses in an effective and scalable way. Using these responses as preferred outputs and treating normal responses as rejected, we perform multiple training rounds to obtain a sequence of models $\mathcal{M}_1, ..., \mathcal{M}_T$, where each model $\mathcal{M}_{t+1}$ utilizes preference data $\mathcal{P}_t$ generated by the
$t$-th model.  Through the information-assisted DPO and iterative training strategy, we progressively endow \modelname with strong demo-driven video in-context learning capabilities.
\section{Experiments}
We first perform detailed evaluation results and analyses on \benchname in \cref{sec:videomimic}. We then compare our method against state-of-the-art Video MLLMs on widely used video benchmarks in \cref{sec:common}. Finally, we present key ablation results in \cref{sec:ablation}. These experimental results systematically validate the value of \benchname and the advantages of our framework.

\begin{table*}[t]
\centering
\caption{\textbf{General Video Understanding.} \modelname achieves superior performance on both general temporal understanding and knowledge acquisition tasks, highlighting the effectiveness of the proposed demo-driven video in-context learning framework. These results also demonstrate the robustness and generalizability of our training strategy across diverse evaluation settings.}
\vspace{-6pt}
\small
\adjustbox{width=\linewidth}{ 
\label{tab:main_table}
\renewcommand{\arraystretch}{1.}
\setlength{\tabcolsep}{6pt}
\begin{tabular}{L{140pt}C{30pt}C{80pt}C{30pt}C{80pt}C{30pt}C{40pt}C{40pt}}
\toprule
\multirow{2}{*}{\textbf{Model}} & \multirow{2}{*}{\textbf{Size}} 
& \multicolumn{4}{c}{\textbf{General Temporal Understanding}} 
& \multicolumn{2}{c}{\textbf{Knowledge}} \\
\cmidrule(lr){3-6} \cmidrule(lr){7-8}
& & {\makecell{\textbf{VideoMME} \\ \scriptsize \textbf{(wo / w sub)}}} 
  & {\textbf{MVBench}} 
  & {\makecell{\textbf{Long}\\\textbf{VideoBench}}} 
  & {\makecell{\textbf{MLVU} \\}} 
  & {\textbf{VideoMMMU}} \\
\midrule
\multicolumn{8}{l}{\textit{Proprietary Models}} \\
\midrule
GPT-4V~\citep{openai2024gpt4technicalreport} & - & 59.9/63.3 & 43.7 & 49.2 & 59.1 & - \\
GPT-4o~\citep{openai2024gpt4ocard} & - & 71.9/77.2 & - & 66.7 & 66.7 & 61.2 \\
Gemini-1.5-Pro~\citep{geminiteam2024gemini15unlockingmultimodal} & - & 73.2/79.8 & - & 64.0 & - & 70.4 \\
Gemini-2.5-Pro~\citep{comanici2025gemini25pushingfrontier} & - & 84.3/86.9 & - & - & - & 83.6 \\
\midrule
\multicolumn{8}{l}{\textit{Open-Sourced Video MLLMs}} \\
\midrule
VideoLLaMA2~\citep{cheng2024videollama2advancingspatialtemporal} & 7B & 47.9 / 50.3 & 54.6 & 36.0 & - & - \\
LLaVA-OneVision~\citep{li2024llavaonevisioneasyvisualtask} & 7B & 58.2 / 61.5 & 56.7 & 56.3 & 64.7 & 33.9 \\
VideoLLaMA3~\citep{zhang2025videollama3frontiermultimodal} & 7B & 66.2 / 70.3 & 69.7 & 59.8 & 73.0 & 47.0 \\
LLaVA-Video~\citep{zhang2025llavavideovideoinstructiontuning} & 7B & 63.3 / 69.7 & 58.6 & 58.2 & 70.8 & - \\
Qwen2.5-VL~\citep{bai2025qwen25} & 7B & 65.1 / 71.6 & 69.6 & 56.0 & - & 47.4 \\
InternVL3.5~\citep{wang2025internvl35advancingopensourcemultimodal} & 8B & 66.0 / 68.6 & 72.1 & 62.1 & 70.2 & - \\
\midrule
VILA-1.5~\citep{lin2024vilapretrainingvisuallanguage} & 40B & 60.1 / 61.1 & - & - & 56.7 & 34.0 \\
VideoLLaMA2~\citep{cheng2024videollama2advancingspatialtemporal} & 72B & 61.4 / 63.1 & 62.0 & - & - & - \\
LLaVA-OneVision~\citep{li2024llavaonevisioneasyvisualtask} & 72B & 66.2 / 69.5 & 59.4 & 61.3 & 66.4 & 48.3 \\
LLaVA-Video~\citep{zhang2025llavavideovideoinstructiontuning} & 72B & 70.5 / 76.9 & 64.1 & 61.9 & 74.4 & 49.7 \\
Qwen2.5-VL~\citep{bai2025qwen25} & 72B & 73.3 / 79.1 & 70.4 & 60.7 & 74.6 & 60.2 \\
\midrule
\rowcolor{gray!20!white}
Ola-Video (Base) & 7B & 63.0 / 68.9 & 66.9 & 60.7 & 69.1 & 46.2\\
\rowcolor{blue!10!white}
Demo-ICL (SFT) & 7B & 65.6 / 70.2 & 69.4 & 61.6 & 70.4 & 48.8 \\
\rowcolor{blue!20!white}
Demo-ICL & 7B & 65.2 / 69.7 & 68.6 & 61.2 & 70.4 & 52.6 \\
\bottomrule
\end{tabular}
}
\vspace{-12pt}
\end{table*}

\subsection{\benchname}
\label{sec:videomimic}

\noindent \textbf{Setup.} We evaluate both representative proprietary MLLMs and state-of-the-art open-source video MLLMs, reporting performance across three tasks. In addition to standard evaluations, we design experiments on text-demo and video-demo in-context learning tasks, including settings without explicit guidance, in order to better characterize the current capabilities and limitations of MLLMs in demo-driven video in-context learning. For proprietary models, we consider GPT-4o and Gemini-2.5-Pro. For open-source video MLLMs, we benchmark a diverse set of representative models, including InternVL-3~\citep{zhu2025internvl3}, Qwen2-VL~\citep{wang2024qwen2vlenhancingvisionlanguagemodels}, Qwen2.5-VL~\citep{bai2025qwen25}, Ola~\citep{liu2025olapushingfrontiersomnimodal}, and LLaVA-Video~\citep{zhang2024video}. To capture the role of specialized video reasoning, we further include Video-R1~\citep{li2025videochat} and VideoChat-R1~\citep{feng2025video} as baselines for video reasoning models. Finally, to examine the effect of model capacity, we conduct experiments on both Qwen2.5-VL-7B and Qwen2.5-VL-72B. We also incorporate the base model of Demo-ICL as a reference to more clearly demonstrate the effectiveness and advantages of our training strategy.

\noindent \textbf{Text-demo In-context Learning.}
As shown in \cref{tab:iclresult}, models perform poorly without demonstrations, indicating that in-context learning is essential for task success. When text demonstrations are provided, all models improve, demonstrating their ability to integrate task-specific knowledge from in-context text demonstrations. The extent of this improvement, however, strongly depends on model size: small models typically gain less than 10 points, whereas Qwen2.5-VL-72B improves by over 20 points, despite performing no better than smaller models without demonstrations. This highlights model scale as a critical factor for effective in-context learning. On \modelname, the SFT model improves by over 10 points through targeted demonstration strategies, while the DPO model achieves state-of-the-art results among models of similar size. These results confirm that well-designed data curation, combined with preference-based training, substantially enhances generalization and efficiency in video in-context learning.

\noindent \textbf{Video-demo In-context Learning.} In the Video-demo ICL task, performance diverges from text-demo results: while some models extract information from video demonstrations, the gains are limited, and models such as InternVL-3, Qwen2-VL, and Ola even suffer degradation. This highlights the difficulty current MLLMs face in extracting and transferring temporal–visual cues for effective ICL. By contrast, \modelname, equipped with demo-driven video ICL, consistently benefits from video demonstrations, though less pronounced compared to text-demo ICL tasks. Our findings indicate that dedicated strategies for video demonstrations are essential to narrow the gap between text and video guidance and to unlock further multimodal generalization.

\noindent \textbf{Demonstration Selection.} 
To approximate real-world scenarios where models must retrieve relevant demonstrations from large video pools, we evaluate them on the demonstration selection task. This task assesses the ability to identify the correct reference video and answer the corresponding questions. We report both video selection accuracy and final question accuracy conditioned on the selected video. Results show that current models often struggle to capture global semantic information, leading to failures in retrieving appropriate demonstrations and producing a substantial gap from human performance. Existing approaches lack not only effective mechanisms for knowledge extraction and transfer, but also robust search and selection capabilities essential for demo-driven video in-context learning in real-world scenarios. Further analysis is provided in \cref{sec:ablation}.

\subsection{General Video Understanding}
\label{sec:common}
\noindent \textbf{Setup.} To evaluate the generalization ability of our \modelname model, we conduct experiments on several widely used video benchmarks. Our analysis focuses on two main directions. First, we assess video knowledge acquisition using benchmarks such as VideoMMMU~\citep{hu2025videommmuevaluatingknowledgeacquisition}, a representative dataset designed to test how models acquire knowledge from videos. In this setting, the model must watch an entire video and answer questions based on its content, thereby evaluating its ability to learn, retain, and apply information in new contexts. This directly highlights the effectiveness of demo-driven video in-context learning, as the model uses demonstrations to generalize beyond the training distribution. Second, we evaluate on general temporal understanding benchmarks, including VideoMME~\citep{videomme}, MVBench~\citep{li2024mvbench}, LongVideoBench~\citep{wu2024longvideobench}, and MLVU~\citep{zhou2024mlvu}, which target diverse tasks such as common video perception, action recognition, and long video understanding. Together, these benchmarks provide a comprehensive evaluation of \modelname, covering both its demo-driven in-context learning generalization and its foundational video understanding capabilities.

\noindent \textbf{Results.} 
As shown in \cref{tab:main_table}, \modelname demonstrates competitive performance across all open-source MLLMs. On the knowledge acquisition benchmark VideoMMMU, \modelname performs on par with recently released models of comparable size and even surpasses several larger counterparts. These results underscore not only the strong capability of the model in visual reasoning, but also the effectiveness of the demo-driven video in-context learning paradigm for scalable knowledge acquisition. Using demonstrations within the input context, \modelname extends its understanding beyond memorized content, reflecting a step towards more flexible and human-like video comprehension. Furthermore, for general video understanding benchmarks, \modelname exhibits performance comparable to models of similar size, suggesting that the proposed demo-driven ICL mechanism can be seamlessly integrated without compromising general video understanding, while simultaneously enhancing knowledge acquisition. Collectively, these findings indicate that Demo-ICL offers a promising and scalable pathway for advancing video reasoning and general temporal understanding.

\subsection{Analysis Experiments}
\label{sec:ablation}
In this section, we present detailed analyses focusing on the challenges of video-demo in-context learning and strategies for training an effective demo-driven model. We highlight the limitations of current models and demonstrate the effectiveness of information-assisted DPO.

\noindent \textbf{Why is the Video-demo ICL task challenging?} We provide deeper insights into why the Video-demo ICL task poses significant challenges for current MLLMs, with results summarized in \cref{tab:ablation_icl}.
First, we test \modelname with more densely sampled frames, and the improvements demonstrate that fine-grained visual cues are critical for demo-driven video in-context learning. 
\begin{wraptable}{r}{4.5cm}
\vspace{-10pt}
\caption{\label{tab:ablation_icl}\small{Ablation study on evaluation settings.}}
\vspace{-6pt}
\adjustbox{width=\linewidth}{
    \begin{tabular}{l|c} 
        \toprule
        Settings & Video-demo ICL \\
        \midrule
        Base(32 frames) & 29.4\\
        128 frames & 30.4 \\
        +Repeat Video & 38.6\\
        +Reference Clips &  35.8 \\
        +ASR \& Captions & \textbf{45.4}\\
        \bottomrule
    \end{tabular}
}
\vspace{-10pt}
\end{wraptable}
Using 128 frames, we further conduct an experiment where the reference video is identical to the query video, thus providing the model with the full content as context. 
The performance gains in this setting suggest that direct grounding and perception are far easier than knowledge transfer through in-context demonstrations, as the model can process visuals effectively but struggles to adapt that knowledge to new scenarios. We further evaluate the use of reference clips as contextual demonstrations, where only the segments depicting the immediate next-step action are provided as in-context examples. This setting reveals a fundamental challenge for Video-demo ICL: models struggle to accurately align and match temporal evidence across demonstrations. Moreover, replacing clips with ASR transcripts and captions yields additional improvements, revealing that current MLLMs still lack robust fine-grained video comprehension and often fail to abstract or summarize clips into precise knowledge for reasoning and further adaptation.

Taken together, these findings highlight why video-demo ICL is uniquely challenging: it requires not only perception but also temporal alignment, abstraction, and flexible knowledge transfer. This underscores the need for models that can truly leverage demonstrations as dynamic sources of contextual information, a critical capability for advancing video understanding and reasoning.

\noindent \textbf{How to train a good demo-driven video in-context learning model?}
\begin{table}
\caption{\label{tab:ablation_training}\textbf{Training setting ablations.} Instructional Video SFT, Info-Assisted DPO, and iterative DPO training strategies collectively enhance performance on the Demo-ICL-Bench benchmark.}
\vspace{-6pt}
\adjustbox{width=\linewidth}{
    \begin{tabular}{l|cccc} 
        \toprule
        Settings &Text-ICL &Video-ICL & DS & Avg\\
        \midrule
        w/o Instructional Videos & 34.0 & 26.2 & 19.0 & 26.4\\
        \modelname(SFT) & 38.4 & 29.4 & 21.5 & 29.8\\
        Vanilla DPO & 40.0 & 30.0 & 22.0 & 30.7\\
        \modelname(DPO) 1-round & 41.8 & 30.8 & 22.5 & 31.7\\
        \modelname(DPO) & \textbf{43.4} & \textbf{32.0} & \textbf{24.0}  & \textbf{33.1} \\
        \bottomrule
    \end{tabular}
}
\vspace{-16pt}
\end{table}
We perform ablation studies to assess the effectiveness of our training strategies, with the results summarized in \cref{tab:ablation_training}. The findings indicate that incorporating instructional videos allows \modelname to use in-context demonstrations and adapt to novel scenarios, yielding significant improvements on \benchname. Our results emphasize the importance of high-quality instructional data in enabling models to generalize beyond basic perception and toward deeper contextual video understanding.

We further investigate the impact of training algorithms. When trained with vanilla DPO, the model struggles to produce high-quality responses, yielding noisy data pairs and only marginal improvements. In contrast, our information-assisted DPO method provides richer feedback signals, which significantly enhance response quality and overall performance. Through an iterative training strategy, \modelname gradually learns from in-context demonstrations and ultimately reaches superior performance. These comparisons reveal that both the quality of demo-driven video data and the design of training strategies are essential for effective video in-context learning. Together, these results indicate that building a strong video in-context understanding model requires not only carefully structured demonstrations but also training paradigms that encourage the model to leverage contextual information.
\vspace{-3pt}
\section{Conclusion}

In this paper, we introduce a novel task, \taskname, which focuses on learning from in-context instructional demonstrations. To facilitate evaluation, we construct \benchname, a benchmark consisting of 1,200 challenging questions designed to assess demo-driven video in-context learning capabilities. To effectively address this task, we further propose \modelname, a video MLLM equipped with enhanced in-context learning abilities. Extensive experiments reveal that existing MLLMs struggle with \taskabbr, while \modelname demonstrates improved performance, achieving superior video understanding and in-context knowledge acquisition capabilities, suggesting its potential to facilitate future developments in this area.

\appendix

\section*{Appendix}
We provide supplementary documents to support our research. Implementation details are outlined in Section~\ref{sec:implementation}. Additional visualization results are presented in Section~\ref{sec:vis}, followed by further experimental analysis in Section~\ref{sec:exp}. We also provide a more comprehensive discussion of related work in Section~\ref{sec:supp_related}. Finally, we discuss the limitations of our work in Section~\ref{sec:limit}.

\section{Implementation Details}
\label{sec:implementation}
\subsection{Experiment Details}
In this section, we detail the implementation of \modelname. The \modelname model is built upon Ola-Video, a highly pretrained multimodal understanding model that integrates OryxViT as its visual encoder to process native arbitrary-resolution visual inputs, alongside Qwen2.5 as the language model. For the training process, we construct a customized dataset to establish foundational image and video understanding capabilities. For image data, resolutions range from 768 to 1536, while for video data, the number of frames is capped at 64, with frame resolutions varying between 288×288 pixels and 480×480 pixels. During training, the maximum token length is set to 16,384, and a learning rate of 1e-5 is used throughout both stages. In the DPO (Direct Preference Optimization) training phase, we curate 5,000 samples using the specified pipeline and apply a learning rate of 5e-7. A batch size of 256 is maintained across both fine-tuning stages and the DPO phase, with experiments conducted using 64 NVIDIA A100 80G GPUs.
\subsection{Data Collection Details}
In the data generation process, we utilize Qwen2.5-72B as our LLM and Qwen2.5-VL-72B as our MLLM within the pipeline. For generating text instructions, we first use Qwen2.5-72B to create summarized instruction steps. Then, when refining these steps with the MLLM, we forward each step along with 64 uniformly sampled frames from the corresponding video clips. For generating questions for video-demo ICL, we provide the text instructions of paired videos and ask the LLM to assess their reasonableness for question generation. Both the LLM and MLLM are deployed using four NVIDIA A800 GPUs.
\subsection{Evaluation Details}
For evaluation on the \benchname, we design three experimental settings. For the Text-demo ICL task, we sample 32 frames from each video and provide textual demonstrations alongside the question during evaluation. For the Video-demo ICL task, we sample 32 frames for both the reference (demonstration) video and the target video. Finally, for the Demonstration Selection task, due to context-length limitations in some models, we uniformly sample 16 frames from each candidate demonstration video and 32 frames from the target video. In all three settings, video subtitles are not provided.

\section{Visualizations}
\label{sec:vis}
\begin{figure*}
    \centering
    \includegraphics[width=1.\linewidth]{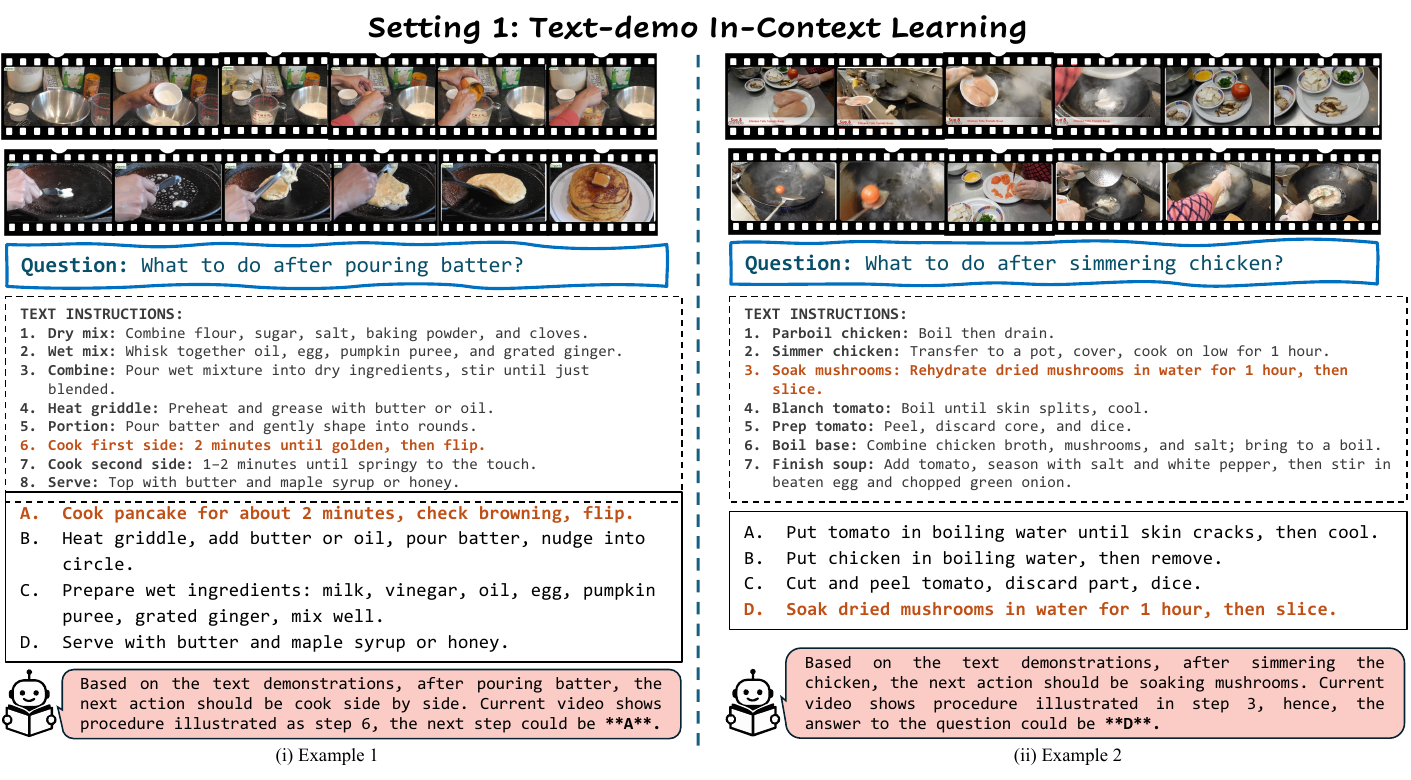}
    \caption{\textbf{Visualization of Text-demo In-Context Learning.} This figure provides 2 examples to illustrate the text-demo in-context learning task, where the text instructions will be provided along with the target video as the inputs.}
    \label{fig:supp_setting1}
\end{figure*}

\begin{figure*}
    \centering
    \includegraphics[width=1.\linewidth]{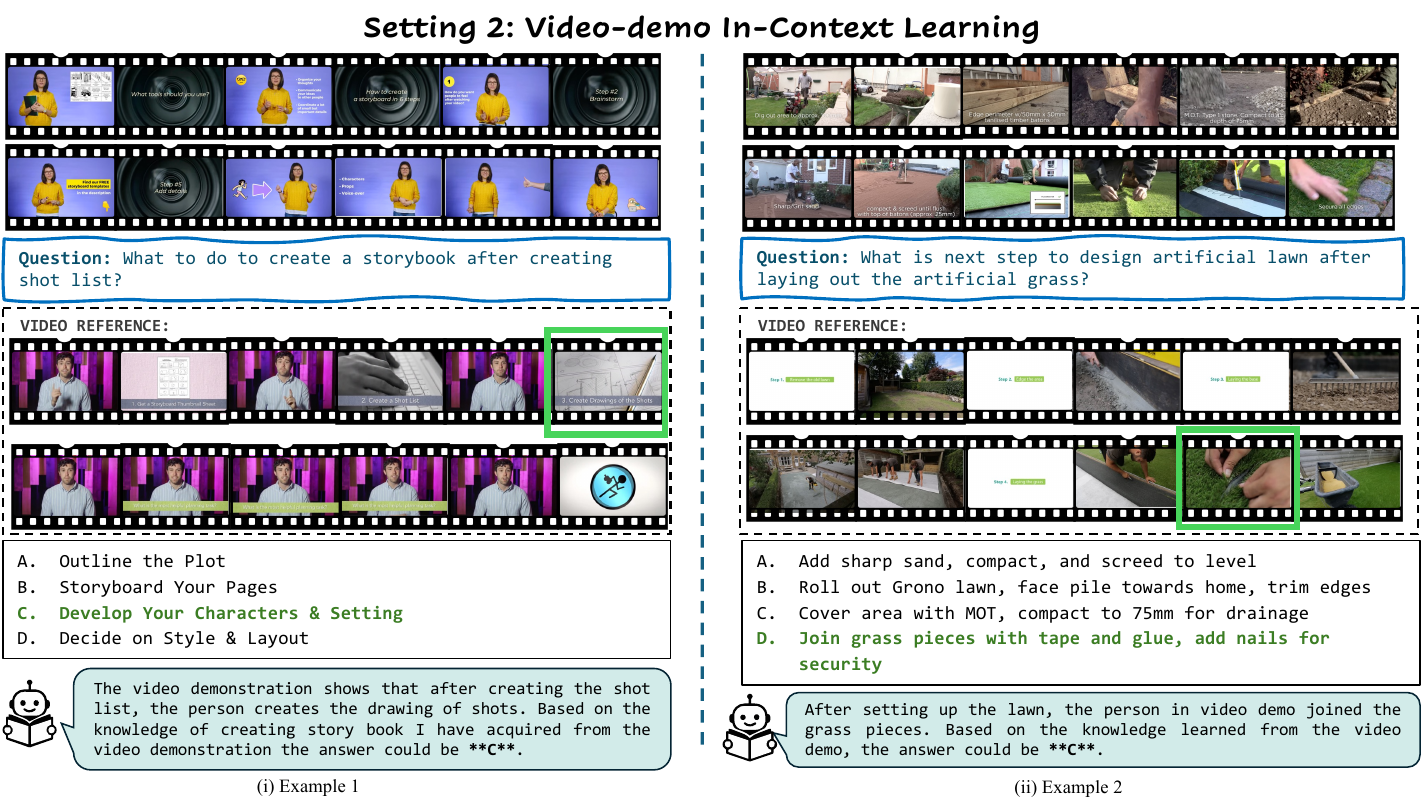}
    \caption{\textbf{Visualization of Video-demo In-Context Learning.} This figure provides 2 examples to illustrate the video-demo in-context learning task, where a video demonstration will be provided together with the target video input.}
    \label{fig:supp_setting2}
\end{figure*}

We present visualization results to clarify the task design of Demo-ICL-Bench. These results are shown in Fig.~\ref{fig:supp_setting1} and Fig.~\ref{fig:supp_setting2}.

\section{More Analysis Experiments}
\label{sec:exp}

\subsection{General Video Understanding on Video-MME}

We further evaluate the Demo-ICL model on general video understanding tasks of varying lengths and scenarios. Specifically, we employ the VideoMME benchmark to highlight its offline video comprehension capabilities, providing a broader assessment beyond domain-specific settings.

\paragraph{Setup.}
To further evaluate the generalization ability of Demo-ICL on diverse video understanding tasks, we adopt the Video-MME benchmark~\citep{videomme}. The dataset consists of 900 videos (254 hours) covering 6 visual domains and 30 subfields, with durations ranging from 11 seconds to 1 hour, categorized into Short, Medium, and Long. In addition to visual content, VideoMME provides audio and subtitles, enabling a multimodal and comprehensive evaluation of video MLLMs. Under this setting, the model is required to watch an entire video and then answer corresponding questions, which allows us to systematically assess robustness across varying durations, modalities, and domains, in comparison with both open-source and commercial MLLMs.

\paragraph{Results.}
Table~\ref{tab:mllm_video_mme} summarizes the overall performance of Demo-ICL across short, medium, and long video tracks. Demo-ICL achieves strong results on all three tracks, demonstrating robust capabilities across different temporal lengths. It surpasses open-source video MLLMs with similar parameter sizes (7B), achieves comparable results to larger models (34B), and competes closely with some commercial MLLMs. Notably, on long-duration videos, which pose greater challenges due to extended temporal dependencies, Demo-ICL demonstrates its long video understanding capabilities, maintaining consistent performance over time.


\begin{table*}[t]
\centering
\caption{\textbf{Performance of Demo-ICL compared to previous MLLMs on Video-MME across short, medium, and long durations, under without ``subtitles'' and with ``subtitles'' settings.}}
\label{tab:mllm_video_mme}
\vspace{-6pt}
\small
\adjustbox{width=0.9\linewidth}{
\renewcommand{\arraystretch}{1.}
\setlength{\tabcolsep}{6pt}
\begin{tabular}{lcccccccccccc}
\toprule
\multirow{2}{*}{\textbf{Models}} & \multirow{2}{*}{\textbf{LLM Params}} & \multicolumn{2}{c}{\textbf{Short (\%)}} & \multicolumn{2}{c}{\textbf{Medium (\%)}} & \multicolumn{2}{c}{\textbf{Long (\%)}} & \multicolumn{2}{c}{\textbf{Overall (\%)}} \\
\cmidrule(lr){3-4} \cmidrule(lr){5-6} \cmidrule(lr){7-8} \cmidrule(lr){9-10}
& & w/o subs & w/ subs & w/o subs & w/ subs & w/o subs & w/ subs & w/o subs & w/ subs \\
\midrule
\multicolumn{10}{c}{\textit{Proprietary MLLMs}} \\
\midrule
GPT-4V~\citep{openai2023gpt4v} & - & 70.5 & 73.2 & 55.8 & 59.7 & 53.5 & 56.9 & 59.9 & 63.3 \\
GPT-4o~\citep{openai2024gpt4o} & - & 80.0 & 82.8 & 70.3 & 76.6 & 65.3 & 72.1 & 71.9 & 77.2 \\
Gemini 1.5 Flash~\citep{gemini2024gemini15} & - & 79.7 & 83.6 & 68.4 & 74.7 & 61.1 & 68.8 & 70.3 & 75.0 \\
Gemini 1.5 Pro~\citep{gemini2024gemini15} & - & 81.7 & 84.5 & 74.3 & 81.0 & 67.4 & 77.4 & 75.0 & 81.3 \\
\midrule
\multicolumn{10}{c}{\textit{Open-source Video MLLMs}} \\
\midrule
LongVA~\citep{zhang2024longcontexttransferlanguage} & 7B & 61.1&	61.6	&50.4	&53.6	&46.2	&47.6 & 52.6&	54.3 \\
VITA 1.5~\citep{fu2025vita15gpt4olevelrealtime} & 7B & 67.0	&69.9	&54.2&	55.7&	47.1	&50.4 & 56.1 & 58.7 \\
mPLUG-Owl3~\citep{ye2024mplug} & 7B & 70.0 & 72.8 & 57.7 & 66.9 & 50.1 & 64.5 & 59.3 & 68.1 \\
TimeMarker~\citep{chen2024timemarker} & 8B & 71.0 & 75.8 & 54.4 & 60.7 & 46.4 & 51.9 & 57.3 & 62.8 \\
MiniCPM-V 2.6~\citep{yao2024minicpm} & 8B & 71.3 & 73.5 & 59.4 & 61.1 & 51.8 & 56.3 & 60.9 & 63.7 \\
\midrule
VILA-1.5~\citep{lin2024vilapretrainingvisuallanguage} & 34B & 68.1 & 68.9 & 58.1 & 57.4 & 50.8 & 52.0 & 59.0 & 59.4 \\
Oryx-1.5~\citep{liu2024oryx}& 34B & 77.3 & 80.6 & 65.3 & 74.3 & 59.3 & 69.9 & 67.3 & 74.9 \\
Qwen2-VL~\citep{wang2024qwen2vlenhancingvisionlanguagemodels}& 72B & 80.1 & 82.2 & 71.3 & 76.8 & 62.2 & 74.3 & 71.2 & 77.8 \\
LLaVA-Video~\citep{zhang2024llavanextvideo} & 72B & 81.4	&82.8	&68.9	&75.6	&61.5	&72.5 & 70.6	&76.9	 \\

\midrule
\rowcolor{blue!20!white}
\modelname & 7B & 78.6 & 79.1  & 63.9 & 68.8 & 53.2 & 61.1 & 65.2 & 69.7 \\
\bottomrule
\end{tabular}
}
\end{table*}

\begin{table}[t]
    \centering
    \caption{\textbf{Video Knowledge Acquisition on Video-MMLU.} \modelname achieves superior performance on the Video-MMLU benchmark, showing substantial gains on the Quiz track, where the model must leverage knowledge from lecture videos to answer novel reasoning questions. These results demonstrate the effectiveness of our demo-driven video in-context learning paradigm and the strong generalization capability enabled by the info-assisted DPO training strategy.}
    \label{tab:videommlu}
    \adjustbox{width=0.48\textwidth}{
    \renewcommand{\arraystretch}{1.05}
    \small
    \begin{tabular}{lcccc}
    \toprule
    \textbf{Model} & \textbf{Size} & \textbf{Notebook} & \textbf{Quiz} & \textbf{Overall} \\
    \midrule
    \multicolumn{5}{c}{\textit{Proprietary MLLMs}} \\
    \midrule
    Gemini-1.5-Flash & - & 39.5 & 47.8 & 43.6 \\
    GPT-4o & - & 53.9 & 44.9 & 49.4 \\
    Claude-3.5-sonnet & - & 67.4 & 71.2 & 69.3 \\
    \midrule
    \multicolumn{5}{c}{\textit{Open-Source MLLMs}} \\
    \midrule
    InstructBLIP-7B & 7B & 19.3 & 2.9 & 11.1 \\
    Cambrian-8B & 8B & 20.2 & 5.2 & 12.7 \\
    LLaVA-1.5 & 7B & 22.3 & 9.1 & 15.7 \\
    Video-LlaVA-7B & 7B & 15.3 & 16.5 & 15.9 \\
    
    LLaVA-NeXT (Vicuna) & 7B & 18.1 & 24.9 & 21.5 \\
    Qwen-VL & 7B & 24.4 & 19.6 & 22.0 \\
    mPLUG-Owl3 & 7B & 22.6 & 22.6 & 22.6 \\

    LLaVA-NeXT (LLaMA3) & 8B & 16.5 & 30.1 & 23.3 \\
    InternVL2-8B & 7B & 31.4 & 16.7 & 24.1 \\
    LLaVA-NeXT (Mistral) & 7B & 20.3 & 31.5 & 25.8 \\
    Qwen2-VL-7B & 7B & 34.2 & 23.4 & 28.8 \\
    LLaVA-NeXT-Video-7B & 7B & 35.8 & 27.4 & 31.6 \\
    LLaVA-OneVision-OV & 7B & 34.6 & 33.4 & 34.0 \\

    Qwen2.5-VL-7B & 7B & 42.0 & 32.9 & 37.5 \\
    InternVL2.5-8B & 8B & 34.5 & 44.5 & 39.5 \\
    \midrule
    \rowcolor{blue!20!white}
    Demo-ICL & 7B & 41.0 & 50.4 & 45.7\\
    \bottomrule
    \end{tabular}

    }
    \
\end{table}

\begin{table}[t]
    \centering
    \caption{\textbf{More experiment results on Demo-ICL-Bench.} The \textbf{T. ICL} and \textbf{V. ICL} refers to the original Text-demo ICL Demo. Acc and Video-demo ICL Demo. Acc, where \textbf{DS} refers to the original DS Acc.}
    \label{tab:moreresult}
    \adjustbox{width=0.48\textwidth}{
    \renewcommand{\arraystretch}{1.05}
    \small
    \begin{tabular}{lccccc}
    \toprule
    \textbf{Model} & \textbf{Size} & \textbf{T. ICL} & \textbf{V. ICL} & \textbf{DS} & \textbf{Overall} \\
    \midrule
    \multicolumn{6}{c}{\textit{Proprietary Methods}} \\
    \midrule
    Gemini-2.5-Pro & - & 54.4 & 36.2 & 26.0 & 38.9 \\
    GPT-4o & - & 48.8 & 31.4 & 24.5 & 34.9 \\
    \midrule
    \multicolumn{6}{c}{\textit{Open-Source Methods}} \\
    \midrule
    Qwen2-VL~\citep{wang2024qwen2vlenhancingvisionlanguagemodels} 
        & 7B & 29.0 & 22.4 & 14.5 & 22.0 \\
    Ola~\citep{liu2025olapushingfrontiersomnimodal} 
        & 7B & 32.2 & 24.6 & 16.0 & 24.3 \\
    LLaVA-Video~\citep{zhang2024video} 
        & 7B & 31.0 & 30.2 & 20.5 & 27.2 \\
    Qwen2.5-VL~\citep{bai2025qwen25} 
        & 7B & 32.8 & 28.0 & 18.0 & 26.3 \\
    InternVL-3~\citep{zhu2025internvl3} 
        & 8B & 31.4 & 27.0 & 16.5 & 25.0 \\
    Video-R1~\citep{feng2025video} 
        & 7B & 33.6 & 27.4 & 17.5 & 26.2 \\
    VideoChat-R1~\citep{li2025videochat} 
        & 7B & 34.4 & 28.2 & 18.5 & 27.0 \\
    \midrule
    \multicolumn{6}{c}{\textit{RAG-based Methods}} \\
    \midrule
    VideoRAG~\citep{ren2025videorag} & - & 52.6 & 22.8 & 18.0 & 31.3\\
    \midrule
    \multicolumn{6}{c}{\textit{Agent-based Methods}} \\
    \midrule
    VideoAgent~\citep{wang2024videoagent} & - & 36.2 & 23.4 & 17.5 & 25.7\\
    \midrule
    \rowcolor{blue!20!white}
    \modelname 
        & 7B & \textbf{43.4} & \textbf{32.0} & \textbf{24.0} & \textbf{33.1} \\
    \bottomrule
    
    \end{tabular}
    }
\end{table}

\begin{table*}[h]
\caption{\textbf{Related Work for Demo-ICL-Bench.} Demo-ICL-Bench stands out due to its demo-driven video in-context learning settings, setting it apart from previous video benchmarks.}\vspace{-5pt}
\label{tab:compare_benchmark}
\centering
\setlength\tabcolsep{15pt}
\resizebox{\textwidth}{!}{
\begin{tabular}{@{}llccccc@{}}
\toprule[1pt]
Benchmark & Video Domain & \#Videos & \#QAs & Video-ICL & Annotation\\ 
\midrule
ActivityNet-QA~\citep{caba2015activitynet} & Human Activities & 800 & 8000 & \textcolor{red}{\ding{55}} & Manual \\
How2QA~\citep{li2020hero} & Instructional Videos & 1166 & 2852 & \textcolor{red}{\ding{55}} & Manual\\
KnowIT-VQA~\citep{garcia2020knowit} & TV Show & 207 & 24k & \textcolor{red}{\ding{55}} & Manual \\
NExT-QA~\citep{xiao2021nextqanextphasequestionansweringexplaining} & Web Videos (Causal/Temporal) & 5.4k & 52k & \textcolor{red}{\ding{55}} & Manual \\
MVBench~\citep{li2024mvbench} & Benchmark Videos & 3641 & 4000 & \textcolor{red}{\ding{55}} & Auto\\
VideoMME~\citep{fu2024video} & YouTube Videos & 900 & 2700 & \textcolor{red}{\ding{55}} & Manual\\
VideoMathQA~\citep{rasheed2025videomathqa} & Instructional Videos & 420 & 420 & \textcolor{red}{\ding{55}} & Manual \\
VideoMMMU~\citep{hu2025videommmuevaluatingknowledgeacquisition} & Lectures & 300 & 900 & \textcolor{red}{\ding{55}} & Manual \\

\midrule
Demo-ICL-Bench & Instructional Videos & 1200 & 1200 & \textcolor{teal}{\checkmark} & Mixed \\
\bottomrule[1pt]
\end{tabular}}
\end{table*}

\subsection{Results on Video-MMLU Benchmark}
To further validate the effectiveness and generalization capability of the proposed demo-driven video in-context learning paradigm and the info-assisted DPO training strategy, we evaluate our approach on the Video-MMLU~\citep{song2025video} benchmark, a video understanding dataset designed to assess lecture comprehension and video-based knowledge acquisition.

\paragraph{Setup.}
The Video-MMLU benchmark consists of 1,065 videos, primarily sourced from lecture recordings. The associated questions are categorized into two types: caption and reasoning questions. Since the answers are open-ended, model performance is assessed using the LLM-as-a-Judge framework, with Qwen-2.5-72B serving as the judge model.

\paragraph{Results.}
As shown in Table~\ref{tab:videommlu}, incorporating demo-driven video in-context learning significantly enhances the model’s ability to learn from lecture videos, where the videos act as demonstrations for subsequent reasoning questions. While the model maintains competitive performance on the caption task, its performance on the Quiz track improves by a substantial margin compared to previous models. It is worth noting that we employ Qwen2.5-7B as the backbone LLM, the same base model used in Qwen2.5-VL, yet our method achieves markedly better results on the Quiz track. This demonstrates the effectiveness of the proposed demo-driven video in-context learning and info-assisted DPO strategy in general video knowledge acquisition.

\subsection{Quality of \benchname}
We conduct a user study to further evaluate the quality of our \benchname. Since human annotators have already verified the quality of the Video-demo ICL tasks, the study primarily focuses on the Text-demo ICL and Demonstration Selection tasks. For the Text-demo ICL task, we examine the accuracy of the textual demonstrations and assess whether the provided information is sufficient for models to correctly answer the corresponding questions. For the Demonstration Selection task, as the correct demonstrations have undergone the same verification process as the Video-demo ICL tasks, we mainly evaluate the overall quality of all candidate video demonstrations. In total, we sample 200 Text-demo ICL tasks and 100 Demonstration Selection tasks, which are manually reviewed by human annotators. The results show that 96\% of the Text-demo ICL tasks pass the human quality check. For the Demonstration Selection tasks, all video samples meet the quality criteria, and in 88\% of the cases, human annotators can confidently identify the correct demonstration video, as shown in ~\ref{tab:main_table} in our paper. We plan to refine and correct the remaining tasks that did not pass the human review to further enhance the overall dataset quality.

\subsection{Why Demonstration Selection Difficult?}
To better understand why the Demonstration Selection task remains challenging for current models, we conduct a detailed analysis of this track.
First, we observe that demonstration selection itself poses significant difficulty. By calculating the demonstration accuracies of existing models, we find that their poor performance indicates an inability to effectively utilize the correct in-context demonstrations when multiple candidates are provided. This suggests that current models exhibit fragile reasoning capabilities in complex, real-world scenarios, where single correct demonstrations can not be directly provided.

Second, based on the Qwen2.5-VL model, we compute the final accuracy for tasks where the model successfully selects the correct demonstrations and compare these results with those from the Video-demo ICL setting for the same questions. Interestingly, even when the correct demonstrations are identified, the model achieves only 22.2\% final accuracy, compared to 25.4\% in the Video-demo ICL setting. These findings reveal that, although current models can recognize the correct videos, they still struggle to focus on relevant information within the selected demonstrations and remain easily distracted by irrelevant context. This analysis also underscores the necessity of info-assisted DPO, which enables models to learn to emphasize the most informative and correct elements within in-context demonstrations, thereby improving overall task performance.

\subsection{The Model Learns or Recalls?}
To assess whether questions in \benchname can be answered using only a model’s internal knowledge, we design a series of validation experiments. First, we evaluate Gemini-2.5-Pro on the original questions without providing any demonstrations or answer options, making the task open-ended. We use Qwen2.5-72B as the LLM judge model. Under this setting, Gemini-2.5-Pro successfully answers only 5\% of the questions. 

These results suggest that \benchname cannot be solved solely through the internal knowledge of current models; they must instead rely on in-context demonstrations to achieve strong performance. Simply recalling prior knowledge is insufficient, as the videos contain detailed, task-specific action sequences on which the questions are based—information that the model cannot infer without actually observing the video content.

\subsection{More Evaluation Results}
To comprehensively evaluate existing video understanding methods on \benchname, we further include additional categories of approaches, specifically RAG-based and agent-based methods. For each category, we select a representative model for evaluation: VideoRAG~\citep{ren2025videorag} for the RAG-based approach and VideoAgent~\citep{wang2024videoagent} for the agent-based approach. As shown in Table~\ref{tab:moreresult}, the RAG-based method VideoRAG performs well on the Text-demo ICL task, primarily because its video knowledge indexing and multimodal retrieval paradigm effectively aligns videos with the provided text demonstrations. However, in the video-demo ICL and demonstration selection tasks, where the model must transfer knowledge learned from demonstrations to solve problems in a target video, VideoRAG still underperforms. This gap underscores the importance of these two task designs and highlights that beyond simple video knowledge acquisition, current video MLLMs must also develop the ability to transfer learned knowledge to new tasks. For the agent-based method VideoAgent, we observe that incorporating agentic strategies enhances performance on the Text-demo ICL task, primarily because they strengthen the grounding capabilities of current video MLLMs and improve alignment between videos and textual demonstrations. However, the method still underperforms on the other tracks, suggesting that while agentic approaches can enhance perceptual accuracy, the ability to transfer learned or perceived knowledge to new tasks remains an open challenge, which is likely crucial for advancing video understanding and reasoning.

\section{More Discussion on Related Works}
\label{sec:supp_related}
In this section, we will include more details of related works.
\paragraph{Multimodal Video Understanding for Knowledge Acquisition.}
Recent research in video understanding has moved beyond low-level perception towards extracting structured knowledge from videos, like procedural steps, events, and concepts. Large-scale instructional datasets have been instrumental in this shift. For example, as mentioned in \ref{sec:related_work}, a lot of instructional datasets~\citep{miech2019howto100m,tang2019coinlargescaledatasetcomprehensive,zhukov2019cross} have driven the development of models that seek to learn high-level knowledge from video, rather than just recognize objects or actions. Moreover, VidSitu~\citep{sadhu2021visualsemanticrolelabeling} addresses video situation recognition by densely annotating 10-second movie clips with semantic role labels, which provides a symbolic knowledge representation of the video. By learning to predict such structured representations, models can acquire a form of event knowledge from videos. Similarly, HT-Step~\citep{afouras2023htstep} aligns the textual instructions from wikiHow~\citep{koupaee2018wikihow} with corresponding segments in instructional videos. It provides 116k temporal segment annotations in 20k how-to videos, each labeled with a step description from wikiHow, enabling models to learn to ground declarative knowledge in procedural video footage.
 
To better learn from such knowledge-intensive data, early multimodal learning approaches applied language-modeling techniques to video data. For example, VideoBERT~\citep{sun2019videobertjointmodelvideo} quantizes video frames into discrete ``visual words'' and then uses a BERT-like transformer to learn joint representations of sequences of visual tokens and narration text. Following models such as ActBERT~\citep{zhu2020actbertlearninggloballocalvideotext} extended this masked language modeling paradigm to action recognition data, and ClipBERT~\citep{lei2021moreclipbertvideoandlanguagelearning} improved efficiency by sampling sparse key frames for end-to-end video-text pretraining. 
By learning from millions of narrated video clips, these models demonstrate an ability to embed procedural and commonsense knowledge implicitly in their representations. \citet{zhou2023procedureawarepretraininginstructionalvideo} proposed the model Paprika used PKG-based pre-trainng procedure to generate psuedo labels for instructional video to train. StepFormer~\citep{dvornik2023stepformerselfsupervisedstepdiscovery} addresses the problem of discovering and localizing key procedure steps in instructional videos without human supervision. It uses video with subtitles (ASR) only, with a transformer decoder that attends to video frames via learnable queries to produce a sequence of key steps. ~\citet{chen2024learninglocalizeactionsinstructional} proposes a framework, MPTVA, that aligns video segments with procedure steps derived via LLM from narration text via long-term semantic similarity and short-term fine-grained similarity. 

\paragraph{Multimodal In-Context Learning.}
Inspired by the textual CoT prompting, recent works curate multimodal datasets with human-written rationales to encourage step-by-step prompting. Video-CoT~\citep{wang2024videocot} provides video QA examples paired with detailed explanations, while Video-Espresso~\citep{han2025videoespresso} scales this approach to large collections of reasoning exemplars. Beyond data-centric methods,~\citet{arnab2025temporalchainthoughtlongvideo} propose Temporal Chain-of-Thought, an inference strategy for long videos where the model iteratively selects relevant clips and reasons over them, enabling efficient multi-step reasoning over extended sequences. A complementary line of work extends retrieval-augmented generation (RAG) to video. VideoRAG~\citep{ren2025videorag} and related work~\citep{tevissen2024towards} index long videos into databases of visual and textual descriptors. At query time, relevant segments and transcripts are retrieved and passed to the language model as context, grounding answers in explicit video evidence. This improves factual accuracy, transparency, and scalability, especially for long videos where direct end-to-end processing is infeasible.

\section{Limitations and Future Directions}
\label{sec:limit}
In this section, we discuss the limitations of our work. The Demo-ICL model does not include a specialized architecture for demo-driven video in-context learning. Instead, we employ a customized training strategy to achieve this functionality. Our goal is to equip current MLLMs with demo-driven video in-context learning capability without requiring architectural modifications, thereby simplifying the integration of these new capabilities and the maintenance of previous multimodal understanding.

Additionally, we did not explore how models can effectively learn from diverse contexts, such as different modalities or resources. This ability is more similar to the natural human learning process, where individuals can draw on a wide range of resources, such as text instructions and instructional videos, to enhance understanding simultaneously. Combining various types of contextual information to improve in-context learning and ultimately enhance a model's performance on new tasks remains a significant challenge.

{
    \small
    \bibliographystyle{ieeenat_fullname}
    \bibliography{cvpr2026}
}

\end{document}